\documentclass[conference]{IEEEtran}
\IEEEoverridecommandlockouts
\usepackage{cite}
\usepackage{amsmath,amssymb,amsfonts}
\usepackage{algorithmic}
\usepackage{algorithm}
\usepackage{graphicx}
\usepackage{textcomp}
\usepackage{xcolor}
\def\BibTeX{{\rm B\kern-.05em{\sc i\kern-.025em b}\kern-.08em
    T\kern-.1667em\lower.7ex\hbox{E}\kern-.125emX}}
    
\makeatletter
\def\ps@IEEEtitlepagestyle{%
\def\@oddfoot{\mycopyrightnotice}%
\def\@evenfoot{}%
}
\def\mycopyrightnotice{%
{\footnotesize 978-1-6654-3288-7/21/\$31.00~\copyright~2021 IEEE\hfill}
\gdef\mycopyrightnotice{}
}    
\begin{document}

\title{
Neural Network Modeling of Probabilities for Coding the Octree Representation of Point Clouds
}

\author{\IEEEauthorblockN{Emre Can Kaya}
\IEEEauthorblockA{\textit{Computing Sciences Unit} \\
\textit{Tampere University}\\
Tampere, Finland \\
emre.kaya@tuni.fi}
\and
\IEEEauthorblockN{ Ioan Tabus}
\IEEEauthorblockA{\textit{Computing Sciences Unit} \\
\textit{Tampere University}\\
Tampere, Finland \\
ioan.tabus@tuni.fi}
}
\maketitle

\begin{abstract}
This paper describes a novel lossless point cloud compression algorithm that uses a neural network for estimating the coding probabilities for the occupancy status of voxels, depending on wide three dimensional contexts around the voxel to be encoded. The point cloud is represented as an octree, with each resolution layer being sequentially encoded and decoded using arithmetic coding, starting from the lowest resolution, until the final resolution is reached. The occupancy probability of each voxel of the splitting pattern at each node of the octree is modeled by a neural network, having at its input the already encoded occupancy status of several octree nodes (belonging to the past and current resolutions), corresponding to a 3D context surrounding the node to be encoded. The algorithm has a fast and a slow version, the fast version selecting differently several voxels of the context, which allows an increased parallelization by sending larger batches of templates to be estimated by the neural network, at both encoder and decoder. The proposed algorithms  yield state-of-the-art results on benchmark datasets. The implementation will be available at https://github.com/marmus12/nnctx
\end{abstract}


\section{Introduction}
\label{sec:intro}

In the recent years, point cloud compression  (PCC) has became a very active field of study, in an effort to provide efficient coding solutions for the very large point clouds available nowadays.  The major contributions to the area are the standardization projects  initiated by JPEG \cite{ebrahimi2016jpeg} and MPEG\cite{schwarz2018emerging}, from which the video PCC (V-PCC) and the geometry PCC (G-PCC) standards are already finalized. Meanwhile, several contributions have appeared in the technical literature, showing improvements over the standardized solutions, for some specific classes of point clouds. Encoding the geometry of voxelized point clouds is a first task, solved both in V-PCC and G-PCC, and for which several recent publications provided alternative solutions, see e.g.: the lossless codec using dyadic decomposition \cite{dd}; the bounding volumes by depthmap projections BVL\cite{bvl}; and more recently, VoxelDNN \cite{voxeldnn} and MSVoxelDNN \cite{msvoxeldnn}, based on deep neural networks for providing the arithmetic coder with coding probabilities.

We propose a neural network based lossless coder for voxelized point clouds. In the algorithm, the internal representation for point clouds is selected to be the octree model, offering multiresolution reconstructions, where at resolution $r$ all points are given with a precision of $r$ bits per dimension. In the octree model each point from the resolution $r-1$ is split into 8 candidate points at resolution $r$, and specifying for each of these 8 points the occupancy status by 1 bit, one can retrieve all points at resolution $r$. Iteratively in the same way, one obtains all resolutions up to the final resolution $R$. Hence, the octree representation recursively constructs the set of points, by specifying an octet for each node at the current resolution. What needs to be encoded for transmitting the point cloud is the octet that represents the splitting pattern at each node, for each resolution level, starting from resolution $0$ up to resolution $r-1$.

Encoding of the splitting octet can be done in several ways. For example, G-PCC encodes the splitting (octet) pattern $ [b_1,b_2,\ldots,b_8]$ at a voxel $n_k$ at resolution $r-1$, by considering for each bit $b_i$ a context defined based on the following: the occupancy status of the neighbors of the current voxel $n_k$ to be splitted,  the location $i$ of the bit $b_i$ inside the octet, the values of the already encoded bits $b_1,\ldots,b_{i-1}$, and  some already encoded splitting patterns at some neighbor voxels of $n_k$. These contexts are further merged, e.g., based on rotation invariance, in order to obtain the most relevant probability models at each context. In here we do not use the context construction of G-PCC, specifically we do not use the probability distribution $p(b_i|b_1,\ldots,b_{i-1}, C)$ conditional at some collapsed context $C$, but instead we associate each $b_i$ to the corresponding voxel at resolution $r$, and establish the context $\cal C$  in the same way, uniformly for all $i$, based on spatially neighbor voxels in the resolution $r-1$ and $r$, as we explain below.  This has the effect of using a single conditional distribution $p(b_i|{\cal C})$  instead of 8 distinct conditional distributions.

VoxelDNN \cite{voxeldnn}, a recent DNN-based method,  generated probabilities for the occupancy of voxels,  using a wider conditional template than in the octree model, by splitting the space in very large cubes, e.g. $64\times64\times64$, and generating conditional probabilities $p(b_i|b_1,\ldots,b_{i-1})$ for each of $i=1,\ldots,64^3$, hence for each voxel inside the large cube. Each such specific distribution leads to an equivalent template with variable 3D shape, handled during the training process by masked convolutions. The equivalent template becomes asymmetrical with respect to  the voxel $b_i$ to be encoded, especially near the facets of the $64\times64\times64$ cube.  In our approach we keep the template of the context the same for most voxels to be encoded, with asymmetries in the template occurring only at the boundary of the overall bounding cube for the point cloud.

\section{Proposed Method}
\label{sec:proposed}

\begin{figure}[tbp]

  \centerline{\includegraphics[width=\columnwidth]{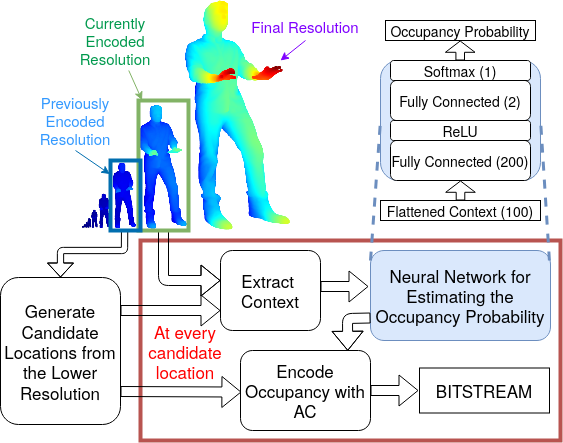}}


%
\caption{The proposed lossless encoding scheme.}
\label{fig:prop}
\end{figure}

The proposed method, which we dub NNOC, is illustrated in Fig. \ref{fig:prop}. When encoding a point cloud with $r$ bits resolution, the encoder encodes additionally all the $(r-1)$ lower resolution versions. Initially, the 2 bits resolution 
point cloud, having $4\times4\times4$ voxels, is simply encoded in 64 bits where each bit represents the occupancy of a voxel.  
Then, the encoding and decoding proceeds by inferring, from the voxels of the point cloud at 2 bits resolution, the possibly occupied (candidate) locations in 3 bits resolution. In general, for every voxel in a resolution level $r-1$, there are 8 possibly occupied voxels in resolution $r$. Hence, if the 2 bits resolution point cloud has $n_p$ points, there are $8n_p$ candidate locations for the 3 bits resolution. The locations other than the candidate locations are known to be unoccupied.



After marking all candidate voxels at resolution $r$ starting from the known voxels at resolution $r-1$, the encoder and decoder proceed to encoding the occupancy status of these candidate voxels, by scanning the candidate locations in a given scanning order (explained below), and at every candidate location a context is constructed, identically at encoder and decoder, encompassing voxels from resolution $r$ that are either candidates or that have their occupancy status known (since they were already scanned). The scanning order of the candidate pixels at resolution $r$ is defined by considering the regular lexicographic one, considering one-by-one the sections across the point cloud at planes $z=z_0$, and in each section scanning is done row-wise (interpreting the section as an image with row index x and column index y). The scanning order establishes the causality status for the voxels in the considered contexts. 
A neural network (NN) model is employed to estimate the probability of occupancy of a candidate voxel given the voxel's context. Then, the occupancy of the candidate voxel is encoded by 2-symbol arithmetic coding using the probability distribution generated by the network. The procedure described above is repeated for all the resolutions starting from 3 bits octree depth until the final resolution of the input point cloud is encoded. Apart from the occupancies of voxels, a necessary side information is the binary occupancy status of each section, stating whether the section contains any points. This is expressed in a binary vector having each element associated to one section, which is encoded with run-length encoding.    
\subsection{Collecting the contexts of candidate voxels}
\label{sec:collcon}
\begin{figure}[tbp]

  \centerline{\includegraphics[width=0.7\columnwidth]{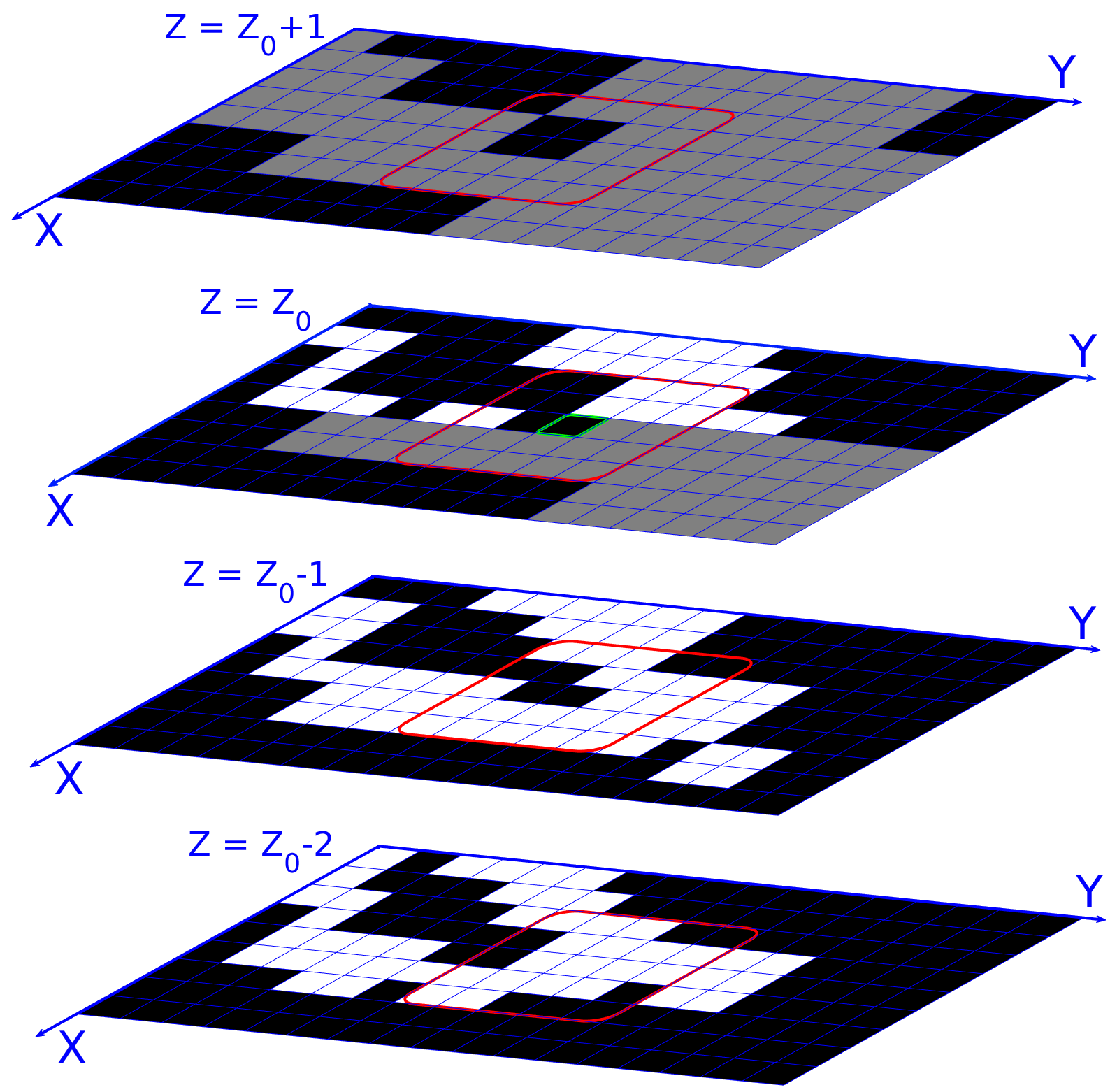}}

\caption{Context definition in the proposed NNOC. Black locations are unoccupied voxels and those that fall into the red bordered windows are represented each by a 0 in the context vector. White and gray locations are occupied and candidate voxels, respectively. Each of those that are contained in the red bordered windows are represented by a 1 in the context vector. The currently encoded/decoded voxel is shown with a green border. }
\label{fig:nnoc}
\end{figure}

Our approach is based on the octree structure and hence we need to encode each bit of the splitting pattern at each node. What first distinguishes  our approach is that at each resolution $r-1$, we traverse the existing nodes and encode their splitting patterns in a two-pass manner as explained next. Suppose we have all nodes at resolution $r-1$, and need to transmit the splitting patterns to reconstruct resolution $r$. For each node $(x_0^{[r-1]},y_0^{[r-1]},z_0^{[r-1]})$, we have 8 possibly occupied children voxels at resolution $r$, namely $(2x_0^{[r-1]}+\alpha,2y_0^{[r-1]}+\beta,2z_0^{[r-1]}+\gamma)$, with $\alpha,\beta,\gamma\in\{0,1\}$, and we store all these points in a sorted list (in the sorting, key $z$ has the highest priority). Having the candidate voxels sorted by section enables us to fetch the section points efficiently. We then reconstruct the points at resolution $r$, plane by plane, by keeping a 4 sections buffer (two already encoded sections, the current one, and the next one not encoded yet) to which we associate 4 binary images as depicted in Fig. \ref{fig:nnoc}. After the status of all candidates in the section $z=z_0$ is encoded, the buffer slides upwards by 1 section, to encode the occupancies of the section $z=z_0+1$.

In terms of the octree coding, what is peculiar in our approach is that we transmit at each resolution $r-1$ the occupancy patterns of the nodes lying in the plane  $ z=z_0^{[r-1]}$, but in two phases: First we transmit the 4 bits of the splitting pattern having $\gamma=0$ (corresponding to the lower plane in the resolution $r$), and after we finish transmitting all these half-splitting patterns for all nodes, we continue transmitting the remaining half-splitting patterns, having $\gamma=1$ (corresponding to the higher plane in the resolution $r$). The process is repeated for the plane $ z=z_0^{[r-1]}+1$. In G-PCC for instance, the scanning order is different, since one transmits at once the eight bits corresponding to the occupancy of the children of $(x_0^{[r-1]},y_0^{[r-1]},z_0^{[r-1]})$, and only then the process moves to a different node at resolution $r-1$.

In the four sections of the buffer, we mark the current state in the reconstruction process: For the past two sections we know the true occupancy status of all the candidate points, so we mark in the past two sections the true occupancy of voxels at resolution $r$. In the next section $z= z_0+1$, we do not yet know any occupancy status, but we know the status of being a candidate, which is marked on the binary image of the section.

We therefore have organized in a convenient way the scanning order for traversing the candidate voxels at each resolution level, in the following order: section-by-section (traversing all voxels lying on a section $z=z_0$), and in each such 2D section taking the scanning order to be row-by-row. Using such a scanning order, the current candidate $(x_0,y_0,z_0)$ whose occupancy needs to be encoded, has some of its 3D neighbors already encoded (they are at those voxels belonging to sections $z=z_0-1$ and $z=z_0-2$, and from the section $z=z_0$ the voxels having the row index $x<x_0$, and finally those voxels having $x=x_0$,  $y<y_0$ and $z=z_0$). The other neighbor voxels either are not candidates (they are not children of some existing point in the resolution $r-1$), or are known to be unoccupied or they are candidates but their occupancy is not known yet.

We define the context as a cuboid with $5\times5\times4$ 3D neighbors of $(x_0,y_0,z_0)$ at resolution $r$ as shown in Fig. \ref{fig:nnoc}. We note that this cuboid changes its position at each encoding of a candidate occupancy, because we place it such that it keeps its alignment with the current candidate.  Due to the selected scanning order, the status of being already encoded or not for each pixel remains the same, so the contexts’ elements keep the same type of information or significance. For example, the neighbor $(x_0-2,y_0-2,z_0-2)$ has always its occupancy known (either because it was not a candidate for testing i.e., not being a child of an existing node at $r-1$), or because it was a candidate and we have transmitted its occupancy status. So the state of $(x_0-2,y_0-2,z_0-2)$  can be either 0, if it is not occupied, or 1 if it is occupied. On the other hand, $(x_0,y_0+2,z_0)$ is not encoded yet. We already can know its occupancy, if it was not a candidate then for sure it is not occupied) but if it was a candidate, its occupancy is not known yet. So the state that can be associated to $(x_0,y_0+2,z_0)$ is 1 if it is  a candidate, and 0 if it is not. 

To conclude, in NNOC we consider a 100-element binary context  vector ${\cal C}$, where each element corresponds to the binary status of one location in the  $5\times5\times4$ template and this status means different things for different voxels: For the already encoded voxels it means occupancy, for not yet encoded voxels it means candidacy. 

Due to practical reasons, the context vector contains the candidacy of also the currently encoded voxel $(x_0,y_0,z_0)$ which is always 1. It is experimentally found that accessing a block of voxels in the current section altogether and writing them to context vector is running faster than going through the positions inside the block one by one to exclude the currently encoded position. 

The significance of the neighbors within the context remains the same for all contexts, and for all resolutions. So we decide to define the probability distribution for the voxel $(x_0,y_0,z_0)$ being occupied ($O(x_0,y_0,z_0) = 1$, or not, $O(x_0,y_0,z_0) = 0$,  $p(O(x_0,y_0,z_0) = 1) = NN({\cal C})$ and propose to implement this function using a neural network, having parameters obtained in a training process. The structure of the algorithm is presented in Algorithm 1.
\begin{algorithm}[t]
\label{algo1}
\caption{Encoding with NNOC }
\begin{algorithmic} 
\REQUIRE A point cloud ${\cal P}_R$ with resolution $R$ bits/dimension

\STATE 1. Construct lower resolution point clouds ${\cal P}_{R-1},\ldots,{\cal P}_2$ representing the nodes at octree depth level $R-1,\ldots,2$.

\STATE 2. Encode ${\cal P}_2$ in 64 bits
\STATE 3. Encode iteratively ${\cal P}_3$ to ${\cal P}_R$ as follows:
\FOR{ $r=3,\ldots,R$} 

\STATE 3.1 Generate for each point $P\in {\cal P}_{r-1}$ the eight candidate voxels in the resolution $r$, resulting in the set of candidate points ${\cal P}_{r}^{C}$

\STATE   3.2 Traverse candidates ${\cal P}_{r}^{C}$ section-by-section and encode occupancies as follows:

\FOR{ $z_0 = 0,\ldots,2^r-1$}

\STATE  3.2.1 Construct 4 binary images as in Fig. 2

\STATE 3.2.2 Traverse the candidates for which $z=z_0$ as described in Section II.A.
\FORALL {$(x_c,y_c,z_0) \in {\cal P}_{r}^{C}$ }
\STATE3.2.2.1   Extract the context vector $\cal C$ from 4 images

\STATE  3.2.2.2  Obtain the coding distribution $NN(\cal C)$

\STATE  3.2.2.3 Encode the occupancy $O(x_c,y_c,z_0)$ using $NN(\cal C)$

\STATE   3.2.2.4 Save the true occupancy  in the image $z=z_0$ 
\ENDFOR
\ENDFOR
\ENDFOR
\end{algorithmic}
\end{algorithm}

\subsection{Generating the coding distribution by a NN having the context $\cal C$ at its input}

The input to the neural network is a binary vector consisting of occupancies or candidancies (as explained in II.A) in the causal context of the location being encoded/decoded. We employ a 3 dimensional context such that the number of inputs elements to the NN is $n_C = 5\times5\times4 = 100$. In order to ensure a reasonable encoding and decoding speed, the neural network structure is kept simple, adopting here a Multilayer Perceptron consisting of 2 fully connected layers. The first layer has $2n_C=200$ neurons with ReLU activations and the 2nd layer has 2 neurons, with outputs $\alpha_1$ and $\alpha_2$, and finally there is a softmax activation giving as output 
\begin{equation}
p(O(x_0,y_0,z_0)= 1)=\frac{e^{\alpha_1}}{e^{\alpha_1}+e^{\alpha_2}}.
\label{eq0}
\end{equation}

The output of the neural network is interpreted as an estimation of the probability distribution of occupancy of the current location given its causal context. This floating point estimation provided by the network is multiplied by $2^{14}$ and rounded to yield integer counts which are used in arithmetic coding. 

The training set for the neural network consists of causal contexts that have occurred at least once in the point clouds selected for training. Let $no_{0i}$ and $no_{1i}$ denote the number of occurences of the $i$'th context in a training batch of contexts (including $b.size$ such contexts), where the current "true" location's occupancy was 0 or 1, respectively. Let $p_{1i}=p(O(x_0,y_0,z_0)=1)$ be the output of the NN and $p_{0i}=1-p_{1i}$. The neural network is trained to minimize the following criterion: 

\begin{equation}
Loss=   -\sum_{i=1}^{b.size} no_{0i}\log_2{p_{0i}} +no_{1i}\log_2{p_{1i}}. \label{eq}
\end{equation}

Note that, if the batch would contain all the causal contexts with corresponding number of occurrences in one point cloud, $Loss$ would be equal to the codelength obtained when compressing that point cloud with arithmetic coding using the estimated probabilities $p_{0i}$ and $p_{1i}$. Therefore, $Loss$ reflects our goal to minimize the codelength in the most natural way. Although the loss formulation looks quite similar to the commonly used cross entropy loss function that is encountered in supervised classification schemes, there are important distinctions worth pointing out: The contexts that occur in the training set are only a small portion of the set of all possible contexts. However, what is expected from the network is to generate a probability distribution for any input context, whether seen or not. Another distinction is that the number of occurrences shouldn't be perceived as a ground truth information. They are obtained from a number of point clouds which are likely to have different number of occurrences than the input point cloud to be compressed. 

The network is trained with batches formed by randomly selected contexts possibly coming from different training point clouds.

\subsection{Parallelization and the fast model}

Parallelization can be utilized efficiently at the encoder, where all the contexts at one section $z=z_0$, can be collected and fed to the network to obtain the probability distributions for all section candidates at once. For high resolutions, this may result in efficient computation of probabilities in GPU in large batches.

\begin{figure}[tbp]

  \centerline{\includegraphics[width=0.7\columnwidth]{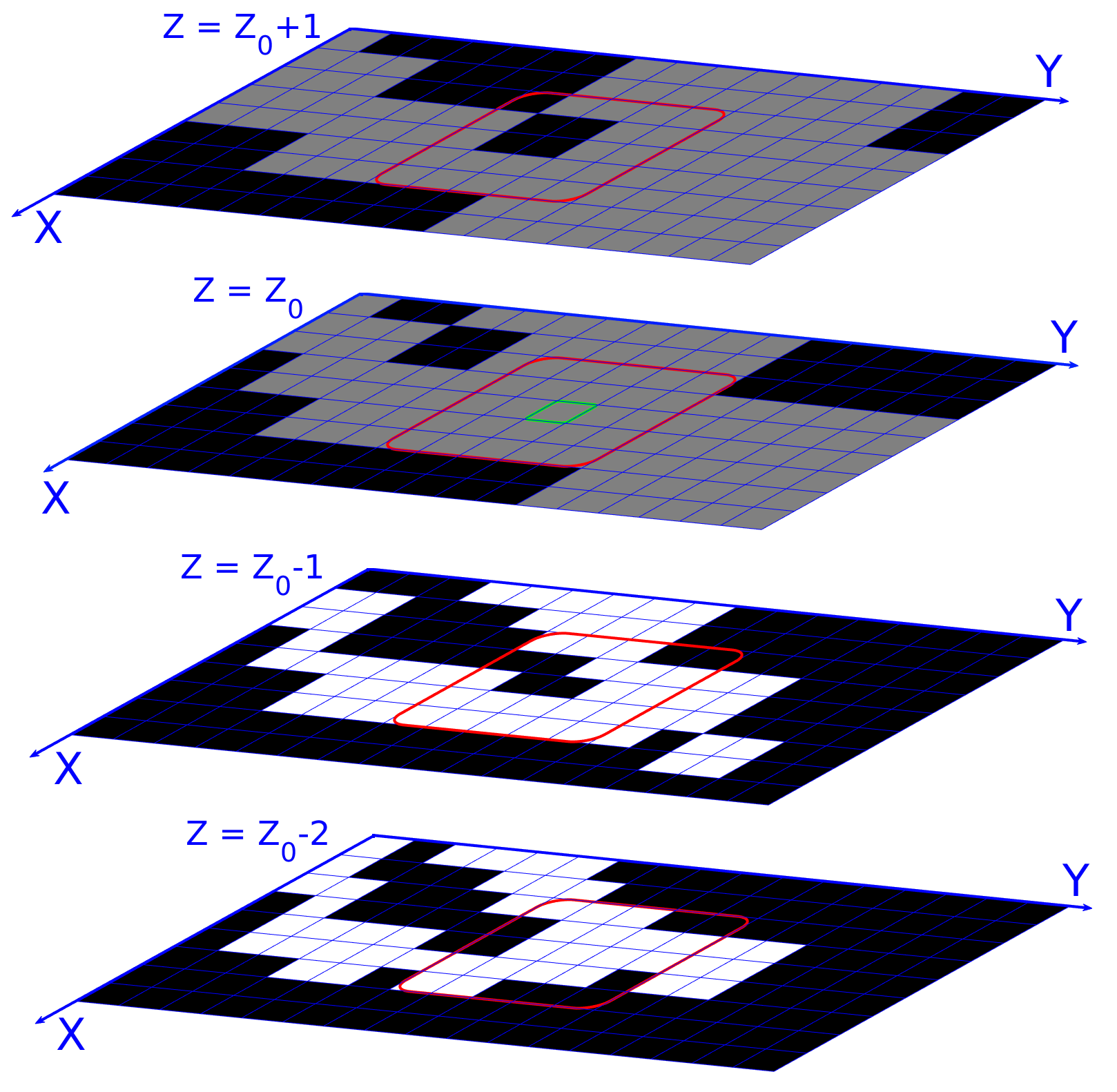}}

\caption{Context definition in fNNOC (the faster version of the proposed NNOC). Black locations are unoccupied voxels and those that fall into red bordered windows are represented each by a 0 in the context vector. White and gray locations are occupied and candidate voxels, respectively. Each of those that are contained in the red bordered windows are represented by a 1 in the context vector. Currently encoded/decoded voxel is denoted with a green window.  }
\label{fig:fnnoc}
\end{figure}

However, at the decoder the situation is different. NNOC requires for forming the context at $(x_0,y_0,z_0)$ the knowledge about the occupancy of $(x_0,y_0-1,z_0)$, which needs to be decoded by arithmetic coding. The only possible sequencing of operations at the decoder is: The decoding of the occupancy of $(x_0,y_0-1,z_0)$ is done and it is used to define the context  ${\cal C}(x_0,y_0,z_0)$. This context is fed to the network which generates the coding distribution of $p(O(x_0,y_0,z_0) = 1) = NN({\cal C})$ that is used to decode the occupancy of $(x_0,y_0,z_0)$ and so on. In this manner, the network operates with batches of size 1, which is simply sequential, with no parallelism.
In order to accelerate the process and to be able to utilize batches of contexts at the decoder, we modify the context in NNOC so that it does not need to make use of the current resolution occupancies at section $z=z_0$. This is realized in the faster version of NNOC called fNNOC. The collection of the context in fNNOC is depicted in Fig. \ref{fig:fnnoc}. This new type of context is for sure less informative and fNNOC performs worse than NNOC in terms of bpov. In fNNOC, the decoder collects all candidates from the current section at once, and then it feeds them to the GPU implementation of the NN, in a batch having the size equal to the number of candidate voxels in section $z=z_0$. As a consequence, durations of encoding and decoding with fNNOC are similar. 

\begin{table*}[!t]
\label{tab:NNOCgain}
\caption{Comparing \textbf{VoxelDNN} \cite{voxeldnn} and \textbf{NNOC} (proposed) in terms of Average Rate [bpov] and Gains over \textbf{G-PCC} \cite{GPCC}}
\begin{center}
 \setlength{\tabcolsep}{5pt}
 \begin{tabular}{|c|c||c|c|c||c|c|c|} 

 \hline
& & \multicolumn{3}{|c||}{\textbf{Single Frame}} &  \multicolumn{3}{ |c|}{\textbf{Average over all frames} } \\
\cline{3-8}
\textbf{Point Cloud(s)}  &  \textbf{Number of} & \textbf{G-PCC} &  \textbf{VoxelDNN }& \textbf{Gain over} &   \textbf{G-PCC} & \textbf{NNOC} &  \textbf{Gain over}\\
&\textbf{ frames}  & & &  \textbf{G-PCC} & & & \textbf{G-PCC}\\
 \hline
   \multicolumn{7}{|c|}{Microsoft Voxelized Upper Bodies \cite{upperbodies}}\\\hline
 Phil9 & 245  & 1.2284  & 0.9201 &  25\% &  1.1785 &  0.8095 &    \textbf{31\%} \\ 
 \hline
  Phil10 & 245 & 1.1617 &   0.8307 &    28\% &  1.135  &    0.7817 &   \textbf{31\%}\\ 
 \hline
 Ricardo9 &  216 & 1.0422 &   0.7173 &  31\% & 1.0836 &  0.6805 &  \textbf{37\%} \\
 \hline 

 Ricardo10 & 216 & 1.0672 &   0.7533 &  29\% &   1.0723 &   0.7006  &   \textbf{35\%} \\
 \hline
\textbf{Average }  & - & 1.1248 & 0.8053 & 28\% &  1.1173 & 0.7431 & \textbf{33\%}\\
 \hline
  \hline
     \multicolumn{8}{|c|}{8i Voxelized Full Bodies \cite{8ivoxelized}}\\\hline

   Loot10 & 300 & 0.9524 &  0.6387 & 33\% &  0.9801 & 0.5904&  \textbf{40\%}\\ 
 \hline
    Redandblack10 & 300 & 1.0889 &  0.7317 & 33\% &   1.1047 & 0.723 &  \textbf{35\%}\\
 \hline
    Boxer9 & 1 & 1.0815 &   0.756 &  30\% &  0.9683 & 0.6439 &   \textbf{34\%}\\ 
     \hline
         Boxer10 & 1 & 0.9 &  0.59 & 34\% &  0.9619 &  0.5507 &   \textbf{43\%}\\     
 \hline
    Thaidancer9 & 1 &  1.0677 &  0.8078 & 24\%  & 1.1253 &  0.7309 &  \textbf{35\%}\\ 
     \hline
    Thaidancer10 & 1 & -  & -  & - & 1.0061   & 0.6839 &    32\%\\ 
     \hline

     \textbf{Average} & - & 1.0476 &  0.7334 & 30\% & 1.0244 & 0.6538 &  36\%   \\ 
 \hline

\end{tabular}

\end{center}

\label{tab:fNNOCgain}
\caption{Comparing \textbf{MSVoxelDNN } \cite{msvoxeldnn} and \textbf{fNNOC} (proposed) in terms of Average Rate [bpov] and Gains over \textbf{G-PCC} \cite{GPCC}}
\begin{center}
 \setlength{\tabcolsep}{5pt}
\begin{tabular}{|c|c||c|c|c||c|c|c|} 
 \hline
& & \multicolumn{3}{|c||}{\textbf{Single Frame}} &  \multicolumn{3}{ |c|}{\textbf{Average over all frames} } \\
\cline{3-8}
\textbf{Point Cloud(s)}  & \textbf{Number of} &  \textbf{G-PCC} &  \textbf{MSVoxelDNN }& \textbf{Gain over} &   \textbf{G-PCC} & \textbf{fNNOC} &  \textbf{Gain over}\\
& \textbf{frames}&  & &  \textbf{G-PCC} & & & \textbf{G-PCC}\\
 \hline
   \multicolumn{8}{|c|}{Microsoft Voxelized Upper Bodies \cite{upperbodies}}\\\hline
 Phil9 & 245 & - & - & - & 1.1785 & 0.9974 & 15\%\\  
 \hline
  Phil10 & 245  & 1.1617 & 1.02 & \textbf{12\%} & 1.135 & 1.0206 & 10\%\\  
 \hline
 Ricardo9 & 216 &  - & - & - & 1.0836 & 0.861 & 21\%\\ 
 \hline 
 Ricardo10 & 216 & 1.0672 & 0.95 & 11\% & 1.0723 & 0.941 & \textbf{12\%}\\ 
 \hline
\textbf{Average } & - & 1.1249 & 0.985 & 11\% & 1.1174 & 0.955 & 14\%\\ 
 \hline
  \hline
     \multicolumn{7}{|c|}{8i Voxelized Full Bodies \cite{8ivoxelized}}\\\hline

   Loot10 & 300 & 0.9524 & 0.73 & 23\% & 0.9801 & 0.7427 & \textbf{24\%}\\ 
 \hline
    Redandblack10 &  300 &  1.0889 & 0.87 & 20\% & 1.1047 & 0.8661 & \textbf{22\%}\\ 
 \hline
    Boxer9 & 1 &  - & - & - & 0.9683 & 0.7466 & 23\% \\ 
     \hline
         Boxer10 &  1 & 0.9 & 0.7 & 22\% & 0.9619 & 0.6815 & \textbf{29\%} \\     
 \hline
    Thaidancer9 & 1 &- & - & - & 1.1253 & 0.8672 & 23\% \\ 
     \hline
    Thaidancer10 & 1 & 1.00 & 0.85 & 15\% & 1.0061 & 0.8069 & \textbf{20\%}\\ 
     \hline

     \textbf{Average} & - &  0.9853 & 0.7875 & 20\% & 1.0244 & 0.78517 & 23.5\%  \\ 
 \hline

\end{tabular}

\end{center}

\caption{Average Rate [Bpov] results on point clouds from CAT1A \cite{ctc} }

\begin{center}
 \begin{tabular}{|c|c||c||c|c||c|c|c|c|c|c|} 
 \hline
 & &\textbf{NNOC}  & \multicolumn{2}{|c||}{\textbf{Other codecs}} &  \multicolumn{6}{ |c|}{\textbf{Fast versions of NNOC} } \\
\cline{3-8}
 \hline
  & \textbf{Bitdepth} & 	 &\textbf{G-PCC } &	\textbf{BVL} \cite{bvl}  & \textbf{fNNOC} &	\textbf{fNNOC1}  &	\textbf{{\sc\bf f}NNOC2} &	\textbf{fNNOC3} & \textbf{fNNOC4} & \textbf{fNNOC5}	 \\ 
  \hline
  \textbf{Input context size} & & 100 &  & & 100 & 75 & 50 & 36 & 100 & 100\\
    \hline
  \textbf{$\#$ of neurons by layer} &  &(200,2)  &  & & (200,2)  &  (150,2) & (100,2) & (72,2) & (200,1) & (200,200,2)  \\
 \hline\hline
  basketball player & 11 & \textbf{0.5934} &	 0.885  & 0.852 &	\textit{0.6908} & 0.8924  & 1.2689 & 0.9098 & 0.6945
 & 0.7083 \\ 
 \hline 
  dancer & 11 &  \textbf{ 0.5751}	 &  0.876 & 0.826 & \textit{0.6907} & 0.8874 &	1.2443	&	0.8931 &  0.6936 & 0.7037
 \\
 \hline
  facade 00064 & 11 & \textbf{1.1053}  & 1.1969 & 1.3331 & 1.2216 &  1.3888 &	1.4973 &	1.3102 & \textit{1.2098 }& 1.2291
\\
 \hline
  queen & 10 & \textbf{0.6897} &	 0.7817  & 0.7883 & \textit{0.9196 }& 1.1327 & 	1.4814 & 1.1773 & 0.9309 & 0.9390
\\
 \hline
 redandblack & 10 & \textbf{0.7353}  & 1.1055 &  1.0418 & 0.8854 & 1.107 &	1.514 &	1.1009 & 0.8932 & \textit{0.8738}
\\  
 \hline
  loot & 10 & \textbf{0.5989} &  0.9818 & 0.8991  & 0.7615 & 0.975 & 1.3785 & 0.9798 & 0.7761 & \textit{0.7557}
\\ 
 \hline
   \textbf{Average} &  & \textbf{0.7163}  & 0.971  & 0.957 & \textit{0.8616} & 1.0639 & 1.3974 & 1.0619 & 0.8663  & 0.8683 \\ 
 \hline
\end{tabular}
\end{center}
\end{table*}

\section{EXPERIMENTAL RESULTS}
\label{sec:expwork}

Training and tests are performed on two publicly available datasets: Microsoft voxelized upper bodies (MVUB) \cite{upperbodies} and 8i voxelized full bodies (8i) \cite{8ivoxelized}. The method is implemented in Python and Tensorflow. Arithmetic coding is adapted from an open source Python implementation \cite{nayuki}. The neural network is trained with contexts collected from 18 randomly chosen frames in Andrew10, David10, Sarah10 sequences (6 from each) from MVUB and 18 randomly chosen frames in Longdress and Soldier sequences (9 from each) from 8i. All of the training contexts are collected from the original resolution of the point clouds (10 bits). The training and validation sequences are not used for tests.  Our selection of training data is similar to the recently introduced VoxelDNN \cite{voxeldnn} and its fast version MSVoxelDNN  \cite{msvoxeldnn}. 

For NNOC, the training set contains around 40 million unique contexts with number of occurences ranging from 1 to 278000 whereas for fNNOC, there are 31 million training contexts with number of occurences ranging from 1 to 135000. These are relatively small training sets considering that the number of possible different contexts with 100 elements is about $2^{100}$. During training we also employ validation sets  through which we decide when to stop training. The validation sets consist of randomly chosen frames from Andrew10 and Soldier sequences. Training was performed using ADAM \cite{adam} optimizer with batches of 30k contexts.

Average bpov results of the proposed NNOC and fNNOC on sequences and individual point clouds from MVUB and 8i datasets are presented in Tables I and II. In Tables I and II, we also show the bpov results obtained with G-PCC,
VoxelDNN and MSVoxelDNN. Since the VoxelDNN and MSVoxelDNN bpovs are for single frames from the sequences, they are not fully comparable with the bpovs for the whole sequences. For a better comparison basis, we show the respective gains in bpov over G-PCC. It should be noted that, Boxer and Thaidancer are in fact individual point clouds (single frames) with bitdepths of 12. We present results for their downsampled versions with bitdepths 9 and 10.

In Table III, left side, we compare the average rates obtained for some of the CAT1A \cite{ctc} point clouds with G-PCC, BVL \cite{bvl} which is a recent lossless method, the proposed NNOC and fNNOC. It is observed that NNOC outperforms all the other methods on all point clouds. fNNOC outperforms BVL on 4 out of 6 point clouds and G-PCC on 5 out of 6 point clouds. A future study might look more into better ways to consider point clouds for training, so that more of the typical existing datasets can be encoded in an efficient way.

\begin{table}
\caption{ Durations of encoding and decoding with NNOC and fNNOC }

\begin{center}
\setlength{\tabcolsep}{3pt}
 \begin{tabular}{|c|c|c|c|c|c|} 
  \hline
  \multicolumn{2}{|c|}{} & \multicolumn{2}{|c|}{\textbf{NNOC}} &  \multicolumn{2}{ |c|}{\textbf{fNNOC} } \\
 \hline
  \textbf{P. Cloud} & \textbf{B.depth} &  \textbf{ENC} & \textbf{DEC} &  \textbf{ENC} & \textbf{DEC} 	 \\ 
 \hline\hline
 phil (1st fr.) & 9 & 1m 17s & 8m 53s & 42s & 46s \\
 \hline
   loot (1st fr.)  & 10 & 3m 11s & 19m 31s  & 1m 46s & 1m 53s \\
 \hline
  basket. player & 11 & 10m 37s	& 3hrs 21m 42s   & 5m 44s & 6m 7s\\ 
 \hline

\end{tabular}
\end{center}
\end{table}
\subsection{Ablation Study}

In order to investigate the effects of the selection of context and network structure, we have performed an ablation study where we train 5 fNNOC variants. Variants fNNOC1-3 have smaller input contexts. fNNOC1 has the same context elements as fNNOC, except removing the voxels coming from the future section ($z_0+1$), hence it has 75 context bits. The second variant called fNNOC2 has the same context elements as fNNOC1 except those coming from the section $z_0-2$, so it has a context vector of length 50. The third variant fNNOC3 has the same 4 context sections as in fNNOC but the context window size is $3\times3$ instead of $5\times5$, so it contains $3\times3\times4=36$ context elements in total. fNNOC4 has a single output neuron representing the occupancy probability, followed by sigmoid instead of 2 neurons and softmax. Finally, fNNOC5 has 2 hidden layers instead of 1. The average rates obtained with the variants are presented in Table III (right side). 

Comparing fNNOC, fNNOC1, fNNOC2 and fNNOC3, one can see that the currently selected context $5\times5\times4=100$ is justified, improving consistently and significantly over the less complex versions. Moreover, comparing fNNOC and fNNOC4, it is observed that having 2 output neurons with softmax instead of 1 with sigmoid yields a slightly better performance. Comparing fNNOC and fNNOC5, it is seen that adding one extra hidden layer does not have a major impact on the bitrate performance.

\subsection{Encoding and Decoding Times}

The encoding and decoding times of NNOC and fNNOC for point clouds with  three different bitdepths are presented in Table IV. While the reported times are not comparable with G-PCC, which can encode and decode in the order of seconds, it might be worth emphasizing that the implementation is done in a high-level environment which is not ideal for speed.

\section{CONCLUSIONS}
\label{sec:conc}

We proposed a lossless compression method that utilizes octree representation and neural network estimation of occupancy probability distributions for splitting the octree nodes, based on contexts obtained through combining information from current and past resolutions. The method has a fast and a slow version and is shown to provide much better compression results than G-PCC, however at a complexity cost which is currently very high. The proposed method compares favorably  with other recently proposed neural network solutions. Further study is needed for expanding the solutions to  larger classes of point clouds and for higher ranges of resolutions.


\bibliographystyle{IEEEbib}
\bibliography{refs}

\end{document}